\documentclass[twoside,11pt]{article}

\usepackage{jmlr2e}
\usepackage{array}
\usepackage{listings}
\usepackage{color}

\usepackage{graphicx}
\usepackage{float}
\usepackage{booktabs}
\usepackage{color}
\usepackage{bm}
\usepackage[american]{babel}

\ShortHeadings{ALiPy: Active Learning in Python}{}
\firstpageno{1}

\begin{document}

\title{ALiPy: Active Learning in Python}

\author{\name Ying-Peng Tang \email tangyp@nuaa.edu.cn
       \AND
       \name Guo-Xiang Li \email guoxiangli@nuaa.edu.cn
   	   \AND
   	   \name Sheng-Jun Huang\thanks{Correspondence author} \email huangsj@nuaa.edu.cn\\   	   
   	   \addr College of Computer Science and Technology, Nanjing University of Aeronautics and Astronautics\\
   	   MIIT Key Laboratory of Pattern Analysis and Machine Intelligence\\
   	   Nanjing 211106, China
}

\editor{}

\maketitle

\begin{abstract}
Supervised machine learning methods usually require a large set of labeled examples for model training. However, in many real applications, there are plentiful unlabeled data but limited labeled data; and the acquisition of labels is costly. Active learning (AL) reduces the labeling cost by iteratively selecting the most valuable data to query their labels from the annotator. This article introduces a Python toobox ALiPy\footnote{http://parnec.nuaa.edu.cn/huangsj/alipy} for active learning. ALiPy provides a module based implementation of active learning framework, which allows users to conveniently evaluate, compare and analyze the performance of active learning methods. In the toolbox, multiple options are available for each component of the learning framework, including data process, active selection, label query, results visualization, etc. In addition to the implementations of more than 20 state-of-the-art active learning algorithms, ALiPy also supports users to easily configure and implement their own approaches under different active learning settings, such as AL for multi-label data, AL with noisy annotators, AL with different costs and so on. The toolbox is well-documented and open-source on Github\footnote{https://github.com/NUAA-AL/ALiPy}, and can be easily installed through PyPI.
\end{abstract}



\section{Introduction}
Active learning is a main approach to learning with limited labeled data. It tries to reduce the human efforts on data annotation by actively querying the most important examples (\cite{S10}). 



ALiPy is a Python toolbox for active learning, which is suitable for various users. On one hand, the whole process of active learning has been well implemented. Users can easily perform experiments by several lines of codes to finish the whole process from data pre-process to result visualization. Also, more than 20 commonly used active learning methods have been implemented in the toolbox, providing users many choices. Table \ref{table:strategies} summarizes the main approaches implemented in ALiPy. On the other hand, ALiPy supports users to implement their own ideas about active learning with high freedom. By decomposing the active learning process into multiple components, and correspondingly implementing them with different modules, ALiPy is designed in a low coupling way, and thus let users to freely configure and modify any parts of the active learning. Furthermore, in addition to the traditional active learning setting, ALiPy also supports other novel settings. For example, the data examples could be multi-labeled, the oracle could be noisy, and the annotation could be cost-sensitive.

\begin{table}[h]
	\centering
	\caption{Implemented active learning strategies in different settings.}   \label{table:strategies}
	\begin{tabular}{p{0.35\columnwidth}| p{0.6\columnwidth}}
		\toprule
		\hline
		
		\textbf{AL with Instance Selection} & Uncertainty (\cite{LewisG94}), Query By Committee (\cite{AbeM98}), Expected Error Reduction (\cite{RoyM01}), Random, Graph Density (\cite{EbertFS12}), BMDR (\cite{WangY13})), QUIRE (\cite{HuangJZ10}), LAL (\cite{KonyushkovaSF17}), SPAL (\cite{TH2019})\\
		
		\hline
		\textbf{AL for Multi-Label Data} & AUDI (\cite{HuangZ13}), QUIRE (\cite{HuangJZ14}), MMC (\cite{YangSWC09}), Adaptive (\cite{LiG13}), Random\\

		\hline
		\textbf{AL by Querying Features}& AFASMC (\cite{HuangXXSNC18}), Stability (\cite{ChakrabortyZBPDY13}), Random \\

		\hline
		\textbf{AL with Different Costs}&  HALC (\cite{YanH18}), Random, Cost performance \\

		\hline
		\textbf{AL with Noisy Oracles}& CEAL (\cite{HuangCMZ17}), IEthresh (\cite{DonmezCS09}), Repeated (\cite{ShengPI08}), Random \\

		\hline
		\textbf{AL with Novel Query Types}& AURO (\cite{HuangCZ15}) \\

		\hline
		\textbf{AL for Large Scale Tasks}& Subsampling \\
		
		\hline
		\bottomrule
	\end{tabular}
\end{table}

\section{Modules in ALiPy}

\begin{figure*}[!t]
	\begin{center} 
		\centering
		\includegraphics[width=0.8\textwidth]{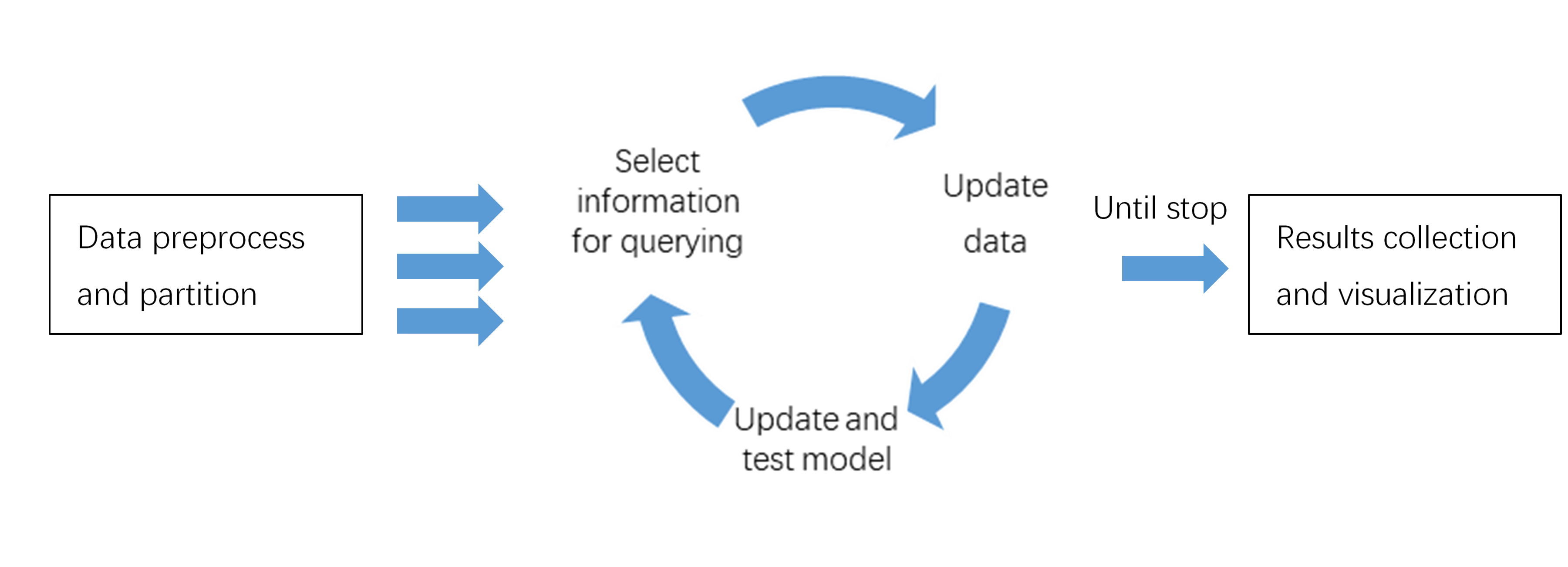}\\
		\caption{A general framework for implementing an active learning approach.} \label{fig:framework}
	\end{center}
\end{figure*}

As illustrated in Figure~\ref{fig:framework}, we decompose the active learning implementation into multiple components. To facilitate the implementation of different active learning methods under different settings, we develop ALiPy based on multiple modules, each corresponding to a component of the active learning process.

Below is the list of modules in ALiPy.
\begin{itemize}
\item \textbf{alipy.data\_manipulate:} It provides the basic functions of data pre-process and partition. Cross validation or hold out test are supported.
\item \textbf{alipy.query\_strategy:} It consists of 25 commonly used query strategies.
\item \textbf{alipy.index.IndexCollection:} It helps to manage the indexes of labeled and unlabeled examples.
\item \textbf{alipy.metric:} It provides multiple criteria to evaluate the model performance.
\item\textbf{alipy.experiment.state} and \textbf{alipy.experiment.state\_io:} They help to save the intermediate results after each query and can recover the program from breakpoints.
\item\textbf{alipy.experiment.stopping\_criteria} It implements some commonly used stopping criteria.
\item\textbf{alipy.oracle:} It supports different oracle settings. One can set to have multiple oracles with noisy annotations and different costs.
\item\textbf{alipy.experiment.experiment\_analyser:} It provides functions for gathering, processing and visualizing the experimental results.
\item\textbf{alipy.utils.multi\_thread:} It provides a parallel implementation of k-fold experiments.
\end{itemize}

The above modules are independently designed implemented. In this way, the code between different parts can be implemented without limitation. Also, each independent module can be replaced by users' own implementation (without inheriting). The modules in ALiPy will not influence each other and thus can be substituted freely. 

In each module, we also provide a high flexibility to make the toolbox adaptive to different settings. For example, in data split function, one can provide the shape of your data matrix or a list of example names to get the split. In the oracle class, one can further specify the cost of each label, and query instance-label pairs in multi-label setting. In the analyser class, the experimental results can also be unaligned for cost-sensitive setting, where an interpolate will be performed automatically when plotting the learning curves.

For more details, please refer to the document at http://parnec.nuaa.edu.cn/huangsj/alipy, and the git repository at https://github.com/NUAA-AL/ALiPy.

\section{Usage of ALiPy}

ALiPy provides several optional usages for different users.

For the users who are less familiar with active learning and want to simply apply a method to a dataset, ALiPy provides a class which has encapsulated various tools and implemented the main loop of active learning, namely \textbf{alipy.experiment.AlExperiment}. Users can run the experiments with only a few lines of codes by this class without any background knowledge.

For the users who want to experimentally evaluate the performance of existing active learning methods, ALiPy provides implementations of more than 20 state-of-the-art methods, along with detailed instructions and plentiful example codes.

For the users who want to implement their own idea and perform active learning experiments, ALiPy provides module based structure to support users to modify any part of active learning. More importantly, some novel settings are supported to make the implementation more convenient. We also provide detailed api references and usage examples for each module and setting to help users get started quickly. Note that, ALiPy does not force users to use any tool classes, they are designed in an independent way and can be substituted by users' own implementation without inheriting anything. 

For details, please refer to the documents and code examples available on the ALiPy homepage and github.

\vskip 0.2in
\bibliography{alipy}

\begin{thebibliography}{20}
\providecommand{\natexlab}[1]{#1}
\providecommand{\url}[1]{\texttt{#1}}
\expandafter\ifx\csname urlstyle\endcsname\relax
  \providecommand{\doi}[1]{doi: #1}\else
  \providecommand{\doi}{doi: \begingroup \urlstyle{rm}\Url}\fi

\bibitem[Abe and Mamitsuka(1998)]{AbeM98}
Naoki Abe and Hiroshi Mamitsuka.
\newblock Query learning strategies using boosting and bagging.
\newblock In \emph{Proceedings of the 15th International Conference on Machine
  Learning}, pages 1--9, 1998.

\bibitem[Chakraborty et~al.(2013)Chakraborty, Zhou, Balasubramanian,
  Panchanathan, Davidson, and Ye]{ChakrabortyZBPDY13}
Shayok Chakraborty, Jiayu Zhou, Vineeth~Nallure Balasubramanian, Sethuraman
  Panchanathan, Ian Davidson, and Jieping Ye.
\newblock Active matrix completion.
\newblock In \emph{{IEEE} 13th International Conference on Data Mining}, pages
  81--90, 2013.

\bibitem[Donmez et~al.(2009)Donmez, Carbonell, and Schneider]{DonmezCS09}
Pinar Donmez, Jaime~G. Carbonell, and Jeff~G. Schneider.
\newblock Efficiently learning the accuracy of labeling sources for selective
  sampling.
\newblock In \emph{Proceedings of the 15th {ACM} {SIGKDD} International
  Conference on Knowledge Discovery and Data Mining}, pages 259--268, 2009.

\bibitem[Ebert et~al.(2012)Ebert, Fritz, and Schiele]{EbertFS12}
Sandra Ebert, Mario Fritz, and Bernt Schiele.
\newblock {RALF:} {A} reinforced active learning formulation for object class
  recognition.
\newblock In \emph{{IEEE} Conference on Computer Vision and Pattern
  Recognition}, pages 3626--3633, 2012.

\bibitem[Huang and Zhou(2013)]{HuangZ13}
Sheng{-}Jun Huang and Zhi{-}Hua Zhou.
\newblock Active query driven by uncertainty and diversity for incremental
  multi-label learning.
\newblock In \emph{{IEEE} 13th International Conference on Data Mining}, pages
  1079--1084, 2013.

\bibitem[Huang et~al.(2010)Huang, Jin, and Zhou]{HuangJZ10}
Sheng{-}Jun Huang, Rong Jin, and Zhi{-}Hua Zhou.
\newblock Active learning by querying informative and representative examples.
\newblock In \emph{Advances in Neural Information Processing Systems}, pages
  892--900, 2010.

\bibitem[Huang et~al.(2014)Huang, Jin, and Zhou]{HuangJZ14}
Sheng{-}Jun Huang, Rong Jin, and Zhi{-}Hua Zhou.
\newblock Active learning by querying informative and representative examples.
\newblock \emph{{IEEE} Transactions on Pattern Analysis and Machine
  Intelligence}, 36\penalty0 (10):\penalty0 1936--1949, 2014.

\bibitem[Huang et~al.(2015)Huang, Chen, and Zhou]{HuangCZ15}
Sheng{-}Jun Huang, Songcan Chen, and Zhi{-}Hua Zhou.
\newblock Multi-label active learning: Query type matters.
\newblock In \emph{Proceedings of the 25th International Joint Conference on
  Artificial Intelligence}, pages 946--952, 2015.

\bibitem[Huang et~al.(2017)Huang, Chen, Mu, and Zhou]{HuangCMZ17}
Sheng{-}Jun Huang, Jia{-}Lve Chen, Xin Mu, and Zhi{-}Hua Zhou.
\newblock Cost-effective active learning from diverse labelers.
\newblock In \emph{Proceedings of the 26th International Joint Conference on
  Artificial Intelligence}, pages 1879--1885, 2017.

\bibitem[Huang et~al.(2018)Huang, Xu, Xie, Sugiyama, Niu, and
  Chen]{HuangXXSNC18}
Sheng{-}Jun Huang, Miao Xu, Ming{-}Kun Xie, Masashi Sugiyama, Gang Niu, and
  Songcan Chen.
\newblock Active feature acquisition with supervised matrix completion.
\newblock In \emph{Proceedings of the 24th {ACM} {SIGKDD} International
  Conference on Knowledge Discovery and Data Mining}, pages 1571--1579, 2018.

\bibitem[Konyushkova et~al.(2017)Konyushkova, Sznitman, and
  Fua]{KonyushkovaSF17}
Ksenia Konyushkova, Raphael Sznitman, and Pascal Fua.
\newblock Learning active learning from data.
\newblock In \emph{Advances in Neural Information Processing Systems}, pages
  4228--4238, 2017.

\bibitem[Lewis and Gale(1994)]{LewisG94}
David~D. Lewis and William~A. Gale.
\newblock A sequential algorithm for training text classifiers.
\newblock In \emph{Proceedings of the 17th Annual International {ACM-SIGIR}
  Conference on Research and Development in Information Retrieval}, pages
  3--12, 1994.

\bibitem[Li and Guo(2013)]{LiG13}
Xin Li and Yuhong Guo.
\newblock Active learning with multi-label {SVM} classification.
\newblock In \emph{Proceedings of the 23rd International Joint Conference on
  Artificial Intelligence}, pages 1479--1485, 2013.

\bibitem[Roy and McCallum(2001)]{RoyM01}
Nicholas Roy and Andrew McCallum.
\newblock Toward optimal active learning through sampling estimation of error
  reduction.
\newblock In \emph{Proceedings of the 18th International Conference on Machine
  Learning}, pages 441--448, 2001.

\bibitem[Settles(2009)]{S10}
B.~Settles.
\newblock Active learning literature survey.
\newblock Technical report, University of Wisconsin-Madison, 2009.

\bibitem[Sheng et~al.(2008)Sheng, Provost, and Ipeirotis]{ShengPI08}
Victor~S. Sheng, Foster~J. Provost, and Panagiotis~G. Ipeirotis.
\newblock Get another label? improving data quality and data mining using
  multiple, noisy labelers.
\newblock In \emph{Proceedings of the 14th {ACM} {SIGKDD} International
  Conference on Knowledge Discovery and Data Mining}, pages 614--622, 2008.

\bibitem[Tang and Huang(2019)]{TH2019}
Ying-Peng Tang and Sheng-Jun Huang.
\newblock Self-paced active learning: Query the right thing at the right time.
\newblock In \emph{Proceedings of the 33rd {AAAI} Conference on Artificial
  Intelligence}, 2019.

\bibitem[Wang and Ye(2013)]{WangY13}
Zheng Wang and Jieping Ye.
\newblock Querying discriminative and representative samples for batch mode
  active learning.
\newblock In \emph{Proceedings of the 19th {ACM} {SIGKDD} International
  Conference on Knowledge Discovery and Data Mining}, pages 158--166, 2013.

\bibitem[Yan and Huang(2018)]{YanH18}
Yifan Yan and Sheng{-}Jun Huang.
\newblock Cost-effective active learning for hierarchical multi-label
  classification.
\newblock In \emph{Proceedings of the 27th International Joint Conference on
  Artificial Intelligence}, pages 2962--2968, 2018.

\bibitem[Yang et~al.(2009)Yang, Sun, Wang, and Chen]{YangSWC09}
Bishan Yang, Jian{-}Tao Sun, Tengjiao Wang, and Zheng Chen.
\newblock Effective multi-label active learning for text classification.
\newblock In \emph{Proceedings of the 15th {ACM} {SIGKDD} International
  Conference on Knowledge Discovery and Data Mining}, pages 917--926, 2009.

\end{thebibliography}

\end{document}